# Learning Universal Multi-level Market Irrationality Factors to Improve Stock Return Forecasting

Chen Yang
School of Computer Science and Engineering
Beihang University
Beijing, China
stx221166@buaa.edu.cn

Jingyuan Wang*
School of Computer Science and Engineering
School of Economics and Management
Beihang University
Beijing, China
jywang@buaa.edu.cn

Xiaohan Jiang
School of Computer Science and Engineering
Beihang University
Beijing, China
jxh199@buaa.edu.cn

Junjie Wu
MIIT Key Laboratory of Data Intelligence and Management, School of Economics and Management
Beihang University
Beijing, China
wujj@buaa.edu.cn

## Abstract

Recent years have witnessed the perfect encounter of deep learning and quantitative trading has achieved great success in stock investment. Numerous deep learning-based models have been developed for forecasting stock returns, leveraging the powerful representation capabilities of neural networks to identify patterns and factors influencing stock prices. These models can effectively capture general patterns in the market, such as stock price trends, volume-price relationships, and time variations. However, the impact of special irrationality factors – such as market sentiment, speculative behavior, market manipulation, and psychological biases – has not been fully considered in existing deep stock forecasting models due to their relative abstraction as well as lack of explicit labels and data description. To fill this gap, we propose *UMI*, a *Universal multi-level Market Irrationality* factor model to enhance stock return forecasting. The *UMI* model learns factors that can reflect irrational behaviors in market from both individual stock and overall market levels. For the stock-level, *UMI* construct an estimated rational price for each stock, which is cointegrated with the stock's actual price. The discrepancy between the actual and the rational prices serves as a factor to indicate stock-level irrational events. Additionally, we define market-level irrational behaviors as anomalous synchronous fluctuations of stocks within a market. Using two self-supervised representation learning tasks, *i.e.,* sub-market comparative learning and market synchronism prediction, the *UMI* model incorporates market-level irrationalities into a market representation vector, which is then used as the market-level irrationality factor. We also developed a forecasting model that captures both temporal and relational dependencies among stocks, accommodating the *UMI* factors. Extensive experiments on U.S. and Chinese stock markets with competitive baselines demonstrate our model's effectiveness and the universality of our factors in improving various forecasting models. We provide our code at https://github.com/lIcIIl/UMI.

## 1 Introduction

Quantitative trading (QT) strategies have seen extensive adoption by financial institutions and investment funds, leading to significant successes in the securities investment markets worldwide. According to Morgan Stanley, quant-based investing has contributed to $1.5 trillion in retail and hedge funds since 2017, and has been particularly favored by the macroeconomic conditions from 2023 to 2024[1]. The stock return forecasting is a crucial technology in QT strategies. It utilizes historical stock features to predict future price trends. In the literature, classic QT strategies adopt specific financial phenomenon to forecast stock returns. For instance, the *momentum* strategy uses the market's momentum to predict stock price trends [17], while the *mean reversion* strategy relies on the principle that asset prices will eventually revert to their average [36].

---

*Corresponding author

[1]https://www.morganstanley.com/ideas/quantitative-investing-outperformance-2023

Recent years have witnessed the revival of deep learning methods [10, 12, 18–22, 33, 37, 38, 46–49, 60], which shed light on developing more effective stock forecasting models. Deep learning models have powerful representation learning capabilities, making them particularly well-suited for handling intricate data relation in real financial markets. Early methods consider stock price or return series as ordinary time series and employ regular sequential deep learning models [1] such as RNN [26], LSTM [50] and Transformer [52]. Subsequently, specially designed neural networks are proposed to handle the characteristics of financial data, such as distribution shifts [61], stochasticity [24], and multiple trading patterns [30]. In addition, many studies have explored incorporating information beyond prices into model inputs to achieve excess returns, such as stock comments [27], experience of trading experts [42], etc.

The core of deep learning-based models lies in using neural networks to extract predictive correlation patterns for stock prices from market behavior data. For existing deep learning stock forecasting models, the correlations they learn typically reflect the most general patterns in the market, such as stock price trend [23], volume-price relationships [25], time variation [9]. However, beyond these general patterns, stock prices are also influenced by many special irrationality factors, such as market sentiment (emotional fluctuations of investors, i.e., panic selling or greedy buying) [43], market manipulation (individual investors or institutions manipulating market prices through improper means) [16], and psychological biases (investor psychological biases such as overconfidence, anchoring effects, loss aversion, etc.) [41]. Compared with the general patterns that can be directly described by explicit labels and side information, the irrationality factors are more abstract and thus challenging to explicitly incorporate into the model. Most existing deep learning forecasting models lack consideration for these irrationality factors, resulting in suboptimal performance.

To bridge this gap, in this paper, we propose a <u>U</u>niversal multi-level <u>M</u>arket <u>I</u>rrationality factor mining model, abbreviated as UMI (You-Me) model. In the UMI model, we categorize irrational behavior in markets into two types: stock-level irrational events and market-level irrational events. A stock-level irrational event is defined as a temporary deviation of a stock's price from its fundamental value (rational price). To exploit these events, we propose a cointegration attention mechanism with stationary regularization to construct an estimated rational price for each stock. This rational price is cointegrated with the actual price (the difference between the actual and rational prices forms a stationary series) but is more stable because it is derived by combining the prices of multiple stocks. The discrepancy between the actual price and the estimated rational price is used as a factor to indicate stock-level irrationalities. On the other hand, we define a market-level irrational event as an anomalous synchronous fluctuation of all stocks in a market, as such broad synchrony in stock fluctuations is atypical in an efficient and rational market. To exploit this type of irrationality, we extract dynamic representations for each stock and combine them into an overall market representation. We then use two self-supervised tasks, i.e., sub-market comparative learning and market synchronism prediction, to incorporate market-level irrationalities into the market representation as market-level irrationality factors. Finally, both stock- and market-level irrationality factors are used as inputs of a deep learning model for stock return forecasting. To fully leverage the irrationality factor, we propose a Transformer and graph-attention mixed model with a RankIC loss as the forecasting model. This model captures correlations in historical stock features and inter-stock relationships, rendering it highly suitable for investment strategies.

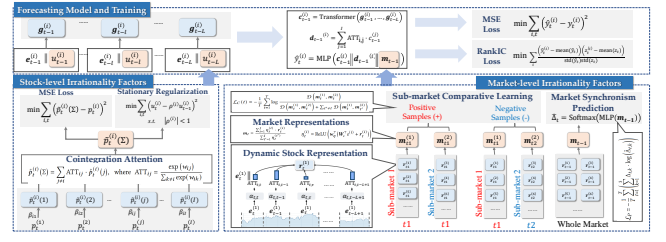

Figure 1: The overview framework of the UMI model.

We have conducted extensive experiments on the stock markets of the U.S. and China, the two largest economies. The results demonstrate the effectiveness of our proposed UMI model in terms of both forecasting accuracy and investment benefits. Notably, the irrationality factors exhibit strong universality in improving various forecasting models. To the best of our knowledge, our work is the first to incorporate market irrational behaviors into deep learning-based stock forecasting.

## 2 Preliminaries

**Basic Notations.** We define a *stock market* as a set of tradable stocks, denoted as $S = \{s_1, \ldots, s_i, \ldots, s_I\}$, where $s_i$ denotes the $i$-th stock. We divide the tradable time of a market as a sequential *periods*, denoted as $\{1, \ldots, t, \ldots, T\}$. Usually, a period is a trading day. For each period, we describe a stock using its features.

DEFINITION 1 (STOCK FEATURE). *The feature of a stock is a set of numerical attributes that describe the state of a stock, such as the opening price, closing price, trading volume,* etc. *The feature of the stock $s_i$ in the $t$-th period is defined as a vector $e_t^{(i)} \in \mathbb{R}^{D_e}$, where $D_e$ is the dimensionality. The feature of $s_i$ during $T$ successive periods is defined as the vector sequence $E^{(i)} = (e_1^{(i)}, \ldots, e_t^{(i)}, \ldots e_T^{(i)})$.*

DEFINITION 2 (STOCK PRICE). *Given $T$ successive periods, the price of the stock $s_i$ is defined as the sequence $p^{(i)} = (p_1^{(i)}, \ldots, p_t^{(i)}, \ldots p_T^{(i)})$, where $p_t^{(i)}$ is the price of $s_i$ at the end of the period $t$. If $t$ is a trading day, $p_t^{(i)}$ is the closing price of that day.*

DEFINITION 3 (STOCK RETURN). *The return of the stock $s_i$ in the period $t$ is defined as*

$$y_t^{(i)} = \left(p_t^{(i)} - p_{t-1}^{(i)}\right) \Big/ p_{t-1}^{(i)} \times 100\%. \tag{1}$$

**Problem Formulation.** The *stock return forecasting* problem is using the historical features and prices of stocks in a market to predict the future price of a stock. Our model adopts a two-step approach to implement a stock return forecasting model. Given the historical features $E_{t-1}^{(i)} = (e_1^{(i)}, \ldots, e_{t-1}^{(i)})$ for the stock $s_i$ in the period $t - 1$, we first adopt a factor learning function to extract indicative factors that can reflect the market irrational behaviors as

$$\Phi_{t-1}^{(i)} = \mathcal{G}\left(\mathcal{E}_{t-1}\right), \tag{2}$$

where $\mathcal{E}_{t-1} = \{E_{t-1}^{(i)}\}_{i=1}^{I}$. $\Phi_t^{(i)}$ are the factors to indicate the influence of market irrationality, namely *irrationality factor Representations*. Next, we employ a stock return forecasting function that takes both $\Phi_{t-1}^{(i)}$ and the historical stock features $\mathcal{E}_{t-1}$ as inputs to predict the future stock return as

$$\hat{y}_t^{(i)} = \mathcal{F}\left(\Phi_{t-1}^{(i)}, \mathcal{E}_{t-1}\right). \quad (3)$$

In the following sections, we provide detailed implementations of the irrationality factor extraction function $\mathcal{G}(\cdot)$ (in Sec. 3 and Sec. 4) and the stock return forecasting function $\mathcal{F}(\cdot)$ (in Sec. 5) for the *UMI* model. The overall framework is illustrated in Fig. 1.

## 3 Stock-level Irrationality Factors

In this section, we propose the method to learn factors that can indicate the market irrational behaviors for individual stocks.

### 3.1 Irrationality of Individual Stocks

**Stock-level Irrational Events.** The stock-level irrational events refer the irrational phenomenon that only creates an investment opportunity for a certain stock. Specifically, we define a stock-level irrational event as the following.

DEFINITION 4 (STOCK-LEVEL IRRATIONAL EVENTS). *Given the real price $p_t^{(i)}$ for the stock $s_i$, we assume there is a rational price $\tilde{p}_t^{(i)}$, if the price divergence between the real price and the rational price is larger than a threshold $H_s$, i.e.,*

$$\left| p_t^{(i)} - \tilde{p}_t^{(i)} \right| > H_s, \quad (4)$$

*there is a stock-level irrational event at the stock $s_i$ in period $t$.*

A temporary deviation of the actual price from the rational price, *i.e.*, an irrational event, creates an investment opportunity. We can profit by buying stocks priced below the rational price or selling stocks priced above it, expecting reversion to the rational price.

**Cointegration Relations.** The challenge in exploiting stock-level irrational events lies in determining the rational price $\tilde{p}_t^{(i)}$. Obviously, $\tilde{p}_t^{(i)}$ is a hidden factor and cannot be directly observed from the historical stock features and prices. Most of the time, we cannot even explicitly calculate rational prices. In classic quantitative trading strategies, one method of addressing this challenge is using the price of a stock closely related to the target stock as a "proxy" for the rational price of the target stock, such prices of the same company's stocks in different markets. The price of a proxy stock is expected to have a *cointegration relations* with the target stock. The definition of cointegration relationships is given in the following [7].

DEFINITION 5 (COINTEGRATION). *Given price series of two stocks, denoted as $\mathbf{p}^{(i)} = (p_1^{(i)}, \ldots, p_T^{(i)})$ and $\mathbf{p}^{(j)} = (p_1^{(j)}, \ldots p_T^{(j)})$, we define the combination of the two series as a new time series,* i.e.,

$$u_t = p_t^{(j)} - \beta p_t^{(i)}, \quad (5)$$

*where $\beta$ is a combination coefficient. If $\mathbf{u} = (u_1, \ldots, u_t, \ldots u_T)$ is a stationary time series, we say $\mathbf{p}^{(i)}$ and $\mathbf{p}^{(j)}$ are* cointegrated.

According to Definition 5, the stationary time series $u_t$ is expected to exhibit fluctuations around its mean value. For a pair of stocks with cointegration relation, if the corresponding $u_t$ deviates significantly from its mean, it indicates the occurrence of an stock-level irrational event. This presents an investment opportunity for us to capitalize on by buying undervalued stocks and selling overvalued ones, *i.e., pair trading*. When $u_t$ reverts to its mean, *i.e.*, the undervalued or overvalued stock return its rational prices, the pair trading strategy yields returns.

**Cointegration Test.** The cointegration relations between two stocks can be identified using the *Dickey–Fuller (DF) Test*. The DF test assumes that the series $\mathbf{u}$ follows an autoregressive process as

$$u_t = \rho u_{t-1} + \varepsilon_t, \quad (6)$$

where $\rho$ is a regression parameter and $\varepsilon_t$ is a random error. The value of $\rho$ determines whether the series $\mathbf{u}$ is stationary. Specifically, we can expand the iterative process of Eq. (6) as

$$\begin{aligned} u_t &= \rho u_{t-1} + \epsilon_t = \rho(\rho u_{t-2} + \epsilon_{t-1}) + \epsilon_t \\ &= \rho(\rho(\rho u_{t-3} + \epsilon_{t-2}) + \epsilon_{t-1}) + \epsilon_t = \ldots \\ &= \rho^{t-1} u_1 + \rho^{t-2} \varepsilon_1 + \rho^{t-3} \varepsilon_2 + \ldots + \varepsilon_t. \end{aligned} \quad (7)$$

As we can see, if $|\rho| >= 1$, the influence of error $\{\varepsilon_t\}_{t=1}^{n}$ to $u_t$ cannot decrease with increasing $t$, leading to non-stationarity. To the contrary, if $|\rho| < 1$, the impact of past errors $\{\epsilon_q\}_{q=1}^{n}$ on $u_t$ diminishes as $t$ increases. This leads to stationarity. The DF test verifies the stationarity of $u_t$ by testing whether $|\rho|$ is less than 1.

### 3.2 Stock-level Irrationality Factor Learning

Although utilizing cointegration to construct a pair trading strategy can capitalize on stock-level market irrationality, its effectiveness remains limited. Natural cointegration relations in the stock market are very sparse. Traditional pair trading strategies require screening all potential stock pairs to identify those that are cointegrated, a method that proves to be highly inefficient. Furthermore, the strategy can only invest in a limited number of cointegrated stocks, resulting in higher investment risk and lower strategy stability. To overcome this limitation, we propose an attention-based method to construct a cointegrated series for each stock as an estimated rational price. This estimated rational price is then used as an indicative factor to enhance the performance of stock return forecasting.

**Cointegration Attention.** Given the price series $p_t^{(i)}$ of stock $i$, we use the price $p_t^{(j)}$ of stock $j \neq i$ to generate a candidate cointegrated price at period $t$ as

$$\tilde{p}_t^{(i)}(j) = \beta_{ij} \cdot p_t^{(j)}, \quad (8)$$

where $\beta_{ij}$ is a learnable parameter. Rather than identifying cointegration relationships between stock $i$ and the candidate cointegrated prices via rigorous testing, we incorporate an attention mechanism to combine all candidate prices into a *virtual rational price* for each $p_t^{(i)}$ as

$$\tilde{p}_t^{(i)}(\Sigma) = \sum_{j \neq i} \text{ATT}_{ij} \cdot \tilde{p}_t^{(i)}(j), \quad \text{where} \quad \text{ATT}_{ij} = \frac{\exp(w_{ij})}{\sum_{k \neq i} \exp(w_{ik})}. \quad (9)$$

$w_{ij}$ are learnable parameters to calculate the attentions $\text{ATT}_{ij}$.

The virtual rational price $\tilde{p}_t^{(i)}(\Sigma)$ is expected to be cointegrated with the real price series of the stock $i$, *i.e.*,

$$u_t^{(i)} = \tilde{p}_t^{(i)}(\Sigma) - p_t^{(i)} \quad (10)$$

should be a stationary time series. To ensure this condition, we introduce a cointegration objective function to train the parameters $\beta_{ij}$ and $w_{ij}$ of the cointegration attention.

**Cointegration Objective Functions.** The cointegration objective function contains a *Regression Loss* and a *Stationary Regularization*. The regression loss is based on Eq. (10). If the time series of $u_t^{(i)}$ in Eq. (10) is a stationary series, the expectation of $\mathbb{E}[u_t^{(i)}]$ should be a constant with the time $t$. Then, Eq. (10) can be considered as using the attention result $\tilde{p}_t^{(i)}(\Sigma)$ to regress stock price $p_t^{(i)}$, where $u_t^{(i)}$ is a regression error, i.e., $p_t^{(i)} = \tilde{p}_t^{(i)}(\Sigma) - u_t^{(i)}$.

Thus, we use a mean square error (MSE) loss to minimize the regression error as

$$\mathcal{L}_\beta = \frac{1}{T \times I} \sum_{t=1}^{T} \sum_{i=1}^{I} \left( p_t^{(i)} - \tilde{p}_t^{(i)}(\Sigma) \right)^2 \tag{11}$$

The stationary regularization is based on the DF test given in Eq. (6). According to Eq. (6), the series of $u_t^{(i)}$ can be considered as an autoregressive process of

$$u_t^{(i)} = \rho^{(i)} u_{t-1}^{(i)} + \varepsilon_t^{(i)}, \tag{12}$$

where $\rho^{(i)}$ is a regression parameter to be estimated. According to DF test, when $|\rho^{(i)}| < 1$, the series $u_t^{(i)}$ is stationary. Therefore, we introduce a constraint $|\rho^{(i)}| < 1$ in the parameter optimization and obtain the stationary regularization as

$$\mathcal{L}_\rho = \frac{1}{T \times I} \sum_{t=1}^{T} \sum_{i=1}^{I} \left( u_t^{(i)} - \rho^{(i)} u_{t-1}^{(i)} \right)^2. \tag{13}$$
$$\text{s.t. } |\rho^{(i)}| < 1, \ \forall i \in \{1, \ldots, I\},$$

Through adjusting the learnable parameters $w_{ij}$ and $\beta_{ij}$, the stationary regularization can make the constructed virtual rational price series $\tilde{p}_t^{(i)}(\Sigma)$ as cointegrated as possible with the real price series $p_t^{(i)}$. Compared to the traditional pair trading strategy, which identifies sparse natural cointegration relationships between actual stock prices to detect stock-level irrational events, our method constructs a "soft" cointegrated series that serves as an estimated rational price for each stock. This approach is more efficient and universally applicable.

We combine the loss function in Eq.(11) with the stationary regularization in Eq.(13) to form the objective function for the estimated irrational price parameters learning as

$$\mathcal{L}_S = \mathcal{L}_\beta(\mathbf{B}, \mathbf{W}_C) + \lambda_1 \mathcal{L}_\rho(\boldsymbol{\rho}), \quad \text{s.t. } |\rho^{(i)}| < 1, \tag{14}$$

where $\lambda_1$ is a hyper-parameter to balance the regression and $i \in \{1, \ldots, I\}$. $\mathbf{B}$ and $\mathbf{W}_C$ are parameter matrices constituting of $\beta_{ij}$ and $w_{ij}$, and $\boldsymbol{\rho}$ is the parameter vector constituting of $\rho^{(i)}$, which is constrained as absolute values less than 1.

Finally, we use $u_t^{(i)} = \tilde{p}_t^{(i)}(\Sigma) - p_t^{(i)}$ as the *stock-level irrationality factor* to indicate local irrational events for the stock $s_i$ in the period $t$. A high value of $u_t^{(i)}$ suggests that the stock $s_i$ is currently undervalued, while a low value of $u_t^{(i)}$ indicates the opposite. We include the stock-level irrationality factor as part of $\Phi_t^{(i)}$ in Eq. (2) and input it into the stock return forecasting function in Eq. (3).

## 4 Market-level Irrationality Factors

### 4.1 Market-level Irrational Events

In a perfectly rational market, the price of an individual stock should be determined solely by its fundamental value and the price trend of rises and falls should not be highly synchronized either. However, in real-world stock markets, the prices of individual stocks are always impacted by overall market sentiment, resulting in the synchronicity of price behaviors [2]. A bullish market sentiment can lead to upward price movement in most stocks, while a bearish one does the opposite. For instance, between Oct. 1973 and Oct. 1974, US stock prices plummeted by 44.1%, whereas from Apr. 1994 to Apr. 1999, they witnessed a remarkable surge of 184.8%. The 2013 Nobel laureate in Economics, Robert J. Shiller posits that such market behaviors are linked to irrationality factors within populations [40]. To quantitatively characterize the irrational behavior of the entire market, we provide the following definition of *Market-level Irrational Events*.

DEFINITION 6 (MARKET-LEVEL IRRATIONAL EVENTS). *Given the stocks $\{s_1, \ldots, s_I\}$ in a market, we denote the increase and decrease of the stocks' returns during period $t$ as $\{\delta_t^{(1)}, \ldots, \delta_t^{(I)}\}$, where $\delta_t^{(i)} = 1$ if the return of stock $s_i$ increases more than a threshold, $\delta_t^{(i)} = -1$ if that decreases more than a threshold, otherwise $\delta_t^{(i)} = 0$. We define that there is a market-level irrational event when the ratio of stocks with $\delta_t^{(i)} = 1$ or with $\delta_t^{(i)} = -1$ are higher than a threshold $H_m$, i.e.,*

$$\left| \sum_{i=1}^{I} \delta_t^{(i)} \right| > H_m. \tag{15}$$

In an efficient and rational market, stock fluctuations typically do not occur in broad synchrony. Therefore, Definition 6 denotes a market-level irrational event as an anomalous synchronous fluctuation of all stocks in a market.

### 4.2 Market-level Irrationality Factor Learning

As Robert J. Shiller emphasizes, the reasons for irrational market behaviors are interconnected with many different factors, such as market structure, culture, and even investor psychology [40]. Simply including market-level irrational events defined in Eq. (15) as factors in a stock return forecasting model cannot adequately capture these underlying factors. Therefore, we adopt a representation learning method to learn comprehensive representation of market-level irrationality. The representation learning comprises two modules: *i)* a market-level representation extraction module for constructing representations that can accurately reflect the behavior of the entire market; and *ii)* two self-supervised tasks aimed at incorporating market-level irrationality into the market representations.

*4.2.1 Market Representation Extraction.* The market representation extraction module first extracts a dynamic behavior representation for each stock, and then combines these stock representations as an overall market representation.

**Dynamic Stock Representation Extraction.** We first extract representation that captures the dynamic behavior of each stock based on its historical features. Specifically given the short-term historical features in a sliding window for a stock, denoted as $(e_{t-L+1}, \ldots, e_{t-1}, e_t)$ [2], where $L$ is a historical slice window. We employ a self-attention mechanism to compute the interdependencies between the stock features in period $t$ and those in historical periods $\tau \in \{t - L + 1, \ldots, t\}$ as

$$\alpha_{t,\tau} = (W_S e_t + b)^\top (W_S e_\tau + b) / \sqrt{D_e}, \quad (16)$$

where $W_S \in \mathbb{R}^{D_e \times D_e}$ and $b \in \mathbb{R}^{D_e}$ are learnable parameters, and $D_e$ is the dimensionality of the feature $e_t$. Next, we use $\alpha_{t,\tau}$ to calculate the attention weight between $t$ and $\tau$ as $\mathrm{ATT}_{t,\tau}$ and employ the weights to combine $e_\tau$ as

$$r_t = e_t \Big\| \sum_{\tau=t-L+1}^{t} \mathrm{ATT}_{t,\tau} \cdot e_\tau, \text{ where } \mathrm{ATT}_{t,\tau} = \frac{\exp(\alpha_{t,\tau})}{\sum_{k=t-L+1}^{t} \exp(\alpha_{t,k})}. \quad (17)$$

Here, $\|$ is the concatenation operator and $r_t$ is the final representation of the stock in the period $t$. The representation $r_t$ not only captures the current stock features $e_t$, but also incorporates temporal dynamic patterns from historical features $e_\tau$, such as trends or volatility. Therefore, we refer to it as a *dynamic stock representation*.

**Market Representation Generation.** In the dynamic stock representations $r_t$, a stock is described exclusively by the dynamic behaviors of its features, implying that two distinct stocks with identical short-term features may yield the same representation. However, in practice, even with the same short-term features, different stocks can influence market sentiment differently due to factors such as being an index component, industry sector, and investor perception. For example, if two companies of the same type have similar short-term historical features, but one is a component of a major index while the other is not, the index component stock will certainly have a greater impact on market sentiment. To capture this phenomenon, we propose a stock ID-involved stock-dependent weight, which is used to combine dynamic stock representations as a market representation.

Specifically, given the $i$-th stock, we utilize one-hot encoding to represent its ID as a binary vector of $\iota^{(i)} \in \{0, 1\}^I$, where the $i$-th element in $\iota^{(i)}$ is 1 and all others are 0. Next, we construct an ID representation matrix as $W_t = (w_t^{(1)}, \ldots, w_t^{(i)}, \ldots, w_t^{(I)}) \in \mathbb{R}^{2D_e \times I}$, where $w_t^{(i)} \in \mathbb{R}^{2D_e}$ is the column vector of the matrix. Using $W_t$ as learnable parameters, we calculate the *ID-involved stock-dependent weight* for the stock $s_i$ as

$$\eta_t^{(i)} = \mathrm{ReLU}\left(w_\eta^\top (W_t^\top \iota^{(i)} + r_t^{(i)})\right), \quad (18)$$

where $\mathrm{ReLU}(\cdot)$ is a Rectified Linear Unit activation function. $w_\eta \in \mathbb{R}^{2D_e}$ is a learnable parameter.

Given the $I$ stocks in the market, we use the stock-dependent weight $\eta_t^{(i)}$ to combine the dynamic stock representations as a market representation for the period $t$:

$$m_t = \frac{\sum_{i=1}^{I} \eta_t^{(i)} \cdot r_t^{(i)}}{\sum_{i'=1}^{I} \eta_t^{(i')}}. \quad (19)$$

Here, we use $\sum_{i'=1}^{I} \eta_t^{(i')}$ to normalize the market representation. Thus, the magnitude of $m_t$ is not affected by the number $I$ of stocks.

---
[2]We omit the superscript $i$ of stock features where it causes no ambiguity.

*4.2.2 Self-supervised Tasks for Market-level Irrationality Factors Learning.* This module use two self-supervised tasks, *i.e.*, *sub-market comparative learning* and *market synchronism prediction*, to incorporate market-level irrationality into the market representation.

**Sub-market Comparative Learning.** According to Definition 6, market-level irrational events are defined as anomalous synchronous fluctuations of stocks in a market. This means that during a market-level irrational event, different parts of the stock market should exhibit similar behaviors. To capture this characteristic in the market representations, we employ a market comparative learning approach to encourage similar representations among different sub-markets during the same time periods.

Specifically, for each time period, we randomly divide the stocks in a market into two sub-markets, labeled as sub-market 1 and sub-market 2. Using Eq. (19), we calculate corresponding market representations for these two sub-markets, denoted as $m_t^{(1)}$ and $m_t^{(2)}$. In the sub-market comparative learning, we consider $m_t^{(1)}$ and $m_{t'}^{(2)}$ as a positive sample if $t = t'$, and as a negative sample if $t \neq t'$. We adopt the InfoNCE loss [35] to ensure that the positive sample pair $m_t^{(1)}$ and $m_t^{(2)}$ is consistent for all periods as

$$\mathcal{L}_C = -\frac{1}{T} \sum_{t=1}^{T} \log \frac{\mathcal{D}\left(m_t^{(1)}, m_t^{(2)}\right)}{\mathcal{D}\left(m_t^{(1)}, m_t^{(2)}\right) + \sum_{t' \neq t} \mathcal{D}\left(m_t^{(1)}, m_{t'}^{(2)}\right)}. \quad (20)$$

Here, $\mathcal{D}(\cdot, \cdot)$ is a criterion function defined as

$$\mathcal{D}\left(m_t^{(1)}, m_{t'}^{(2)}\right) = \exp\left(\frac{1}{|t - t'| + 1}\left(w_M^\top (m_t^{(1)} \| m_{t'}^{(2)}) + b_M\right)\right), \quad (21)$$

where $w_M$ and $b_M$ are learnable parameters. We weight each sample pair using $\frac{1}{|t-t'|+1}$ to ensure that the negative example with a larger time difference $|t - t'|$ contributes to a smaller similarity in the loss function.

**Market Synchronism Prediction.** This task directly uses market representations to predict the synchronism of market defined in Eq. (15). Specifically, we quantify $\sum_i \delta_t^{(i)}$ as a one-hot label $\Delta_t$ with three classes: two classes correspond to market-level irrational events, *i.e.*, $\Delta_t = (1, 0, 0)$ if $\sum_i \delta_t^{(i)} > H_m$, and $\Delta_t = (0, 1, 0)$ if $\sum_i \delta_t^{(i)} < -H_m$; one class for other situations, *i.e.*, $\Delta_t = (0, 0, 1)$ if $H_m \geq \sum_i \delta_t^{(i)} \geq -H_m$. Then, we use the market representation $m_{t-1}$ with a Multilayer Perceptron (MLP) neural network to predict the label $\Delta_t$ as

$$\hat{\Delta}_t = \mathrm{Softmax}(\mathrm{MLP}(m_{t-1})), \quad (22)$$

where $\hat{\Delta}_t$ is the predicted label. We use a cross entropy loss to minimize the error between $\Delta_t$ and $\hat{\Delta}_t$ as

$$\mathcal{L}_P = -\frac{1}{T} \sum_{t=1}^{T} \left(\sum_{k=1}^{3} \Delta_{t,k} \cdot \log\left(\hat{\Delta}_{t,k}\right)\right), \quad (23)$$

where $\Delta_{t,k}$ is the $k$-th element of the vector $\Delta_t$. Finally, the joint loss function for the market-level irrationality factor learning is

$$\mathcal{L}_M = \mathcal{L}_C + \lambda_2 \mathcal{L}_P. \quad (24)$$

Trained by the two self-supervised tasks, the market representation $m_t$ is incorporated with the information of market-level irrationality. We consider $m_t$ as a *Market-level Irrationality Factor*, which is used as a part of $\Phi_t^{(i)}$ in Eq. (2).

## 5 Forecasting Model and Training

The stock-level irrationality factor $u_{t-1}^{(i)}$ and the market-level irrationality factors $m_t$ can be used as inputs to enhance the performance of any stock return forecasting model. Therefore, we name them as universal indicative factors. To fully harness the potential of the universal indicative factors, we propose a hybrid model combining Transformers and graph neural networks to effectively capture the temporal and relation dependencies in stock data.

**Forecasting Model.** Specifically, we concatenate the stock-level factor $u_{t-1}^{(i)}$ and stock features in a historical period window of size $L$ as $g_{t-l}^{(i)} = e_{t-l}^{(i)} \parallel u_{t-l}^{(i)}$, where $l \in \{1, \ldots, L\}$. We then use a Transformer to encode the historical $g_{t-l}^{(i)}$ as

$$c_{t-1}^{(i)} = \text{Transformer}\left(g_{t-L}^{(i)}, \ldots, g_{t-1}^{(i)}\right). \quad (25)$$

Next, we adopt a graph attention-based *Stock Relation Dependency* module to capture the relation among stocks, where the attention between stock $s_i$ and $s_j$ in period $t$ is calculated as

$$\gamma_t^{(i,j)} = \left(W_\gamma \bigl(W_t^\top \iota^{(i)} + r_t^{(i)}\bigr)\right)^\top \left(W_\gamma \bigl(W_t^\top \iota^{(j)} + r_t^{(j)}\bigr)\right), \quad (26)$$

where $W_\gamma \in \mathbb{R}^{2D_e \times 2D_e}$ is a learnable parameter matrix, the vectors $W_t^\top \iota^{(j)}$ and $r_t^{(i)}$ are learned in Eq. (18). We utilize $\gamma_t^{i,j}$ to compute the attention of stock $s_i$ towards other stocks, and leverage these attentions to combine $c_{t-1}^{(i)}$ generated by Eq. (25) as

$$d_{t-1}^{(i)} = \sum_{j=1}^{I} \text{ATT}_{i,j} \cdot c_{t-1}^{(j)}, \quad \text{where} \quad \text{ATT}_{i,j} = \frac{\exp\left(\gamma_t^{(i,j)}\right)}{\sum_{k=1}^{I} \exp\left(\gamma_t^{(i,k)}\right)}. \quad (27)$$

The representation $d_{t-1}^{(i)}$ captures the interdependence between stock $s_i$ and all other stocks $s_j \forall j \neq i$. Finally, the future stock return at period $t$ is predicted using a MLP network as

$$\hat{y}_t^{(i)} = \text{MLP}\left(c_{t-1}^{(i)} \parallel d_{t-1}^{(i)} \parallel m_{t-1}\right), \quad (28)$$

where $m_{t-1}$ is the market-level irrationality factor at period $t-1$, and $\hat{y}_t^{(i)}$ is a forecasted return for the stock $i$.

**RankIC Incorporated Loss Function.** A direct way to optimize the stock return forecasting model is using a mean square error (MSE) loss as

$$\mathcal{L}_{\text{MSE}} = \frac{1}{T \times I} \sum_{t=1}^{T} \sum_{i=1}^{I} \left(\hat{y}_t^{(i)} - y_t^{(i)}\right)^2. \quad (29)$$

In addition to the MSE loss, we also incorporate a *RandIC regularization* to enhance the adaptability of our forecasting model to investment strategies. For many investment strategies, accurate ranking prediction of stock returns holds greater significance than precise numerical prediction of returns. For example, our experiment adopts a *long-short hedging strategy*[3] as the backbone investment strategy for our forecasting model. For each trading period, the long-short hedging strategy first ranks stocks in the market based on their predicted returns. The strategy then takes long positions (buys) in the top-$N$ stocks with the highest predicted returns and short positions (sells) in the bottom-$N$ stocks with the lowest predicted returns. As long as the predicted order of returns is accurate, this hedging strategy can be profitable (Refer to Appendix A.1 for a detailed description of the hedging strategies).

---
[3]https://en.wikipedia.org/wiki/Long/short_equity

We adopt the information correlation (IC) between the predicted stock return and the rank of the true return as a regularization term for the forecasting model training. Specifically, we denote $z_t^{(i)}$ as the rank of the stock $s_i$'s return at period $t$. The IC between $z_t^{(i)}$ and $\hat{y}_t^{(i)}$ is calculated as

$$\text{IC}_t = \frac{1}{I} \sum_{i=1}^{I} \frac{\left(\hat{y}_t^{(i)} - \text{mean}(\hat{y}_t)\right)\left(z_t^{(i)} - \text{mean}(z_t)\right)}{\text{std}(\hat{y}_t)\,\text{std}(z_t)}, \quad (30)$$

where mean($\cdot$) and std($\cdot$) are mean and standard deviation for $\hat{y}_t = (\hat{y}_t^{(1)}, \ldots, \hat{y}_t^{(i)}, \ldots)$ and $z_t = (z_t^{(1)}, \ldots, z_t^{(i)}, \ldots)$ with $i \in \{1, \ldots, I\}$. The $\text{IC}_t$ in Eq. (30) measures the normalized correlation between the predicted return and the rank of true return. For the predicted returns $\hat{y}_t^{(i)}$, even if the prediction accuracy is not very high, as long as their rank is similar to the true return, they can also have a high $\text{IC}_t$.

We define the RankIC regularization for all samples as $\mathcal{L}_{\text{IC}} = -\frac{1}{T} \sum_{t=1}^{T} \text{IC}_t$. The final objective function is defined as

$$\mathcal{L} = \mathcal{L}_{\text{MSE}} + \lambda_3 \mathcal{L}_{\text{IC}}, \quad (31)$$

By incorporating RankIC regularization into the objective function, the model can better capture the relative rank relationship among stocks, which can improve the performance of our model in the hedging investment strategy.

**Factor-learning + Forecasting-training Procedure.** Finally, we implement our model training procedure in two steps. The first step is factor-learning, which uses the training dataset to learn stock-level and market-level irrationality factors by minimizing self-supervised losses $\mathcal{L}_S$ (Eq. (14)) and $\mathcal{L}_M$ (Eq. (24)), respectively. The second step is forecasting-training, which minimizes the objective function $\mathcal{L}$ (Eq. (31)) using supervised labels $y_t^{(i)}$ and $z_t^{(i)}$.

## 6 Experiments
## 6.1 Experiment Setups

**Datasets.** We employ data from the stock markets of the two largest economies: United States (US) and China (CN) to evaluate the performance of our model. The US stock market data contains totally 8993 stocks in American Stock Exchange, New York Stock Exchange and National Association of Securities Dealers Automated Quotations[4] from 03/27/2000 to 10/10/2020, with a test set from 10/10/2015 to 10/10/2020. The CN stock market data contains totally 5148 stocks in Shanghai and Shenzhen Stock Exchanges[5] from 01/10/2006 to 03/01/2023, with a test set from 03/01/2018 to 03/01/2023. We use Alpha360 features of the open-source quantitative investment platform Qlib [56] to extract six price/volume-related features for each stock, which are opening price, closing price, highest price, lowest price, volume-weighted average price, and trading volume. We use the closing price as the stock price to calculate the return. We compare our model to the baselines under *Rolling Retraining* setup,

---
[4]https://github.com/microsoft/qlib/
[5]https://github.com/chenditc/investment_data/

where we retrain the models every 30 trading days. For each retraining, data of new trading days and initial public offerings (IPO) stocks are incorporated into the training sets.

**Model Implementation & Baselines.** For the comparative experiments, seven baseline methods are utilized, including three stock return forecasting methods and four time series forecasting methods.

First, we select two state-of-the-art stock return forecasting models as our baselines

• *DoubleAdapt, (DA)* [61] is a stock return forecasting approach to transform stock data into a locally stationary distribution to handle the impact of distribution shifts in financial data.

• *D-Va* [24] is a stock return forecasting approach that addresses stochasticity in both input and target sequences using a hierarchical variational autoencoder technique.

Second, since the stock RoR prediction can also be considered as a normal time series forecasting problem, we introduce two state-of-the-art time series forecasting methods as baselines. These baselines are equipped with components to handle some special characteristics of complex time series, such as frequency and non-stationarity. These characteristics are also existed in the stock data.

• *SAN* [32] is a time series forecasting approach that introduces a comprehensive normalization framework for non-stationary data.

• *FreTS* [57] is a time series forecasting method that incorporates frequency analysis for forecasting tasks. This approach considers both global perspectives and energy compaction to exploit the frequency information in time series.

Our model employs self-supervised methods to extract hidden factors from stock data. For comparison, we also adopt three state-of-the-art self-supervised methods in our baselines, including a stock return forecasting method and two time series forecasting methods.

• *Co-CPC* [44] is a contrastive learning method in stock market analysis. This approach shuffles the future states of stocks and identifies them as positive or negative samples.

• *Basisformer* [34] is a self-supervised time series forecasting approach that integrates contrastive learning and basis learning for forecasting tasks.

• *SimMTM* [6] is a self-supervised time series forecasting method that utilizes a masked pre-training framework along with the transformer architecture.

**Metrics.** We evaluate the performance of our model from two perspectives: forecasting accuracy and investment strategy performance. For forecasting accuracy, we use RMSE, MAE, IC, ICIR, RankIC and RankICIR as evaluation metrics. The details about the forecasting accuracy metrics are given in Appendix.

For investment strategy performance evaluation, we put the RoR prediction results of our model and baseline into the long-short portfolio hedging investment strategy defined in Eq. (33), and calculate the returns of the investment strategy for each period in the test set. We set the portfolio size $N$ in Eq. (33) to 10% of the total number of stocks per day. To model the effect of transaction costs on the strategy, we adjust the RoR of the portfolio as $R_t = y_t^{STR} - 0.1\% \times \frac{TC}{N}$, where TC is the number of stocks that is transacted at time $t$. Here, the transaction cost of trading a stock is set to 0.1%.

Carrying out trading simulations day by day, we have a portfolio return sequence as $\mathbf{R} = (R_1, \ldots, R_t, \ldots, R_T)$. In the experiment, the investment strategy performance with the return sequence $\mathbf{R}$ is evaluated by five metrics: Annualized Return, Annualized Volatility, Sharpe Ratio, Maximum DrawDown and Calmar Ratio.

• *Annualized Return (AR)*: Annualized Return is a metric that gauges the overall profit generated by investing in line with model predictions on an annual basis. This metric is determined by adjusting the average daily portfolio return to account for the total number of trading days in a year, which is calculated as AR = mean($\mathbf{R}$) × $N_Y$, where $N_Y$ is the total number of trading days in a year.

• *Annualized Volatility (AV)*: Annualized Volatility measures the risk of the portfolio, which is calculated as AV = std($\mathbf{R}$) × $\sqrt{N_Y}$.

• *Sharpe Ratio (SR)*: The Sharpe Ratio is a metric to measure the risk-adjusted return of an investment. It is a measure of how much excess return is generated for a given level of volatility risk. Given the annualized return and annualized volatility, the Sharpe Ratio (SR) is calculated as SR = AR/AV.

• *Maximum DrawDown (MDD)*: The Maximum Drawdown quantifies the worst performance of a portfolio within specific periods. It is calculated as the maximum observed loss that a portfolio experiences during a decline from its highest point to its lowest point. The MDD during $T$ periods is calculated as

$$\text{MDD} = \max_{\tau \in [1,T]} \left( \max_{t \in [1,\tau]} \left( \sum_{\leq t} R_t - \sum_{\leq \tau} R_\tau \right) \right) \quad (32)$$

• *Calmar Ratio (CR)* is the risk-adjusted AR based on Maximum DrawDown. It is calculated as $CR = AR/MDD$.

• *Cumulative Wealth (CW)*: Cumulative wealth is the total accumulation of stock returns over time, reflecting an individual's or entity's net worth resulting from investments. It is calculated as $CW_t = \prod_{t'=1}^{t} (1 + y_{t'}^{(STR)})$.

### 6.2 Forecasting and Investment Performance

**Overall Performance.** Table 1 presents the experimental results of the forecasting and investment performance. Each experiment was repeated five times, and the average results are reported. From the table, we have the following observation. *Firstly*, our model achieves the best performance in both US and CN markets for forecasting and investment performance, demonstrating its effectiveness. *Secondly*, models specifically designed for stock data, such as DoubleAdapt, D-Va, Co-CPC, and our model, demonstrate superior performance compared to general time series forecasting baselines. This suggests that stock series possess unique characteristics that require specialized mechanisms for effective modeling. *Thirdly*, self-supervised methods outperform purely supervised methods. Specifically, Co-CPC surpasses supervised stock forecasting methods, while SimMTM and Basisformer outperform supervised time series forecasting methods. Self-supervised methods can extract hidden information from multiple perspectives through various self-supervised tasks, leading to better performance. Our model employs a self-supervised approach to uncover market irrationality factors that previous methods have overlooked, resulting in significant performance improvements. *Finally*, our model's performance improvements in the rank accuracy (IC/ICIR/RankIC/RankICIR) is much higher than value accuracy (RMSE/MAE), verifying the effectiveness of the RankIC regularization in Eq. (31). For the hedging investment strategy, the relative rank of stock returns is more crucial. Hence, our model's investment performance improvements are significantly greater than its forecasting improvements. Moreover, Fig. 2 compares the cumulative wealth of *UMI* and the baselines. We also provide the cumulative wealth for the market return in Fig. 2. For the CN and US markets, we use CSI500 and S&P500 as their respective market returns. The figure shows that *UMI* significantly outperforms the baselines, with its advantages expanding over time. This indicates that even a small forecasting performance advantage can lead to substantial profit increases in a long-term investment.

Table 1: Comparison of model performance on US and CN markets, with the best results emphasized in bold and the second-best results underlined for clarity. Metrics with (↓) indicate that lower values are better, otherwise the higher values are better.

| | Market | Metrics | DA | D-Va | Co-CPC | SAN | Basisformer | FreTS | SimMTM | *UMI* | Imp. | NR | NS | NM | ND | *UMI* +CM |
|---|---|---|---|---|---|---|---|---|---|---|---|---|---|---|---|---|
| Forecasting Performance | US | RMSE(↓) | .0398 | .0399 | .0401 | .0401 | .0402 | .0400 | .0401 | **.0394** | 1.1% | .0395 | .0397 | .0396 | .0395 | .0393 |
| | | MAE(↓) | .0234 | .0235 | .0234 | .0238 | .0237 | .0239 | .0235 | **.0231** | 1.1% | .0231 | .0232 | .0233 | .0231 | .0230 |
| | | IC(↑) | 0.054 | 0.049 | 0.052 | 0.049 | 0.050 | 0.048 | 0.053 | **0.057** | 6% | 0.055 | 0.055 | 0.056 | 0.056 | 0.061 |
| | | ICIR(↑) | 0.624 | 0.555 | 0.591 | 0.569 | 0.581 | 0.540 | 0.621 | **0.663** | 6% | 0.649 | 0.645 | 0.655 | 0.637 | 0.721 |
| | | RankIC(↑) | 0.057 | 0.055 | 0.055 | 0.053 | 0.053 | 0.050 | 0.055 | **0.061** | 9% | 0.058 | 0.056 | 0.058 | 0.057 | 0.060 |
| | | RankICIR(↑) | 0.615 | 0.606 | 0.603 | 0.588 | 0.598 | 0.553 | 0.604 | **0.666** | 8% | 0.646 | 0.606 | 0.635 | 0.617 | 0.663 |
| | CN | RMSE(↓) | .0354 | .0357 | .0355 | .0360 | .0358 | .0359 | .0358 | **.0350** | 1.1% | .0351 | .0353 | .0352 | .0351 | .0349 |
| | | MAE(↓) | .0225 | .0226 | .0227 | .0229 | .0230 | .0228 | .0227 | **.0222** | 1.4% | .0223 | .0223 | .0225 | .0224 | .0221 |
| | | IC(↑) | 0.069 | 0.071 | 0.071 | 0.065 | 0.068 | 0.062 | 0.068 | **0.078** | 10% | 0.072 | 0.072 | 0.074 | 0.073 | 0.079 |
| | | ICIR(↑) | 0.735 | 0.747 | 0.762 | 0.702 | 0.732 | 0.670 | 0.746 | **0.831** | 9% | 0.773 | 0.775 | 0.785 | 0.791 | 0.828 |
| | | RankIC(↑) | 0.069 | 0.067 | 0.070 | 0.061 | 0.064 | 0.062 | 0.066 | **0.073** | 5% | 0.070 | 0.071 | 0.070 | 0.071 | 0.074 |
| | | RankICIR(↑) | 0.730 | 0.708 | 0.737 | 0.652 | 0.693 | 0.673 | 0.699 | **0.777** | 5% | 0.754 | 0.771 | 0.758 | 0.776 | 0.786 |
| | Market | Metrics | DA | D-Va | Co-CPC | SAN | Basisformer | FreTS | SimMTM | *UMI* | Imp. | NR | NS | NM | ND | *UMI* +CM |
| Investment Performance | US | AR(↑) | 0.220 | 0.177 | 0.213 | 0.154 | 0.172 | 0.142 | 0.180 | **0.289** | 31% | 0.257 | 0.227 | 0.229 | 0.253 | 0.307 |
| | | AV(↓) | 0.174 | 0.157 | 0.156 | 0.160 | 0.155 | 0.156 | 0.165 | **0.144** | 7% | 0.153 | 0.148 | 0.151 | 0.152 | 0.141 |
| | | SR(↑) | 1.260 | 1.134 | 1.366 | 0.971 | 1.119 | 0.901 | 1.100 | **2.007** | 47% | 1.667 | 1.529 | 1.516 | 1.670 | 2.126 |
| | | MDD(↓) | 0.184 | 0.194 | 0.184 | 0.284 | 0.262 | 0.304 | 0.254 | **0.129** | 30% | 0.161 | 0.164 | 0.163 | 0.152 | 0.126 |
| | | CR(↑) | 1.198 | 0.927 | 1.165 | 0.538 | 0.649 | 0.475 | 0.711 | **2.236** | 87% | 1.593 | 1.386 | 1.402 | 1.664 | 2.434 |
| | CN | AR(↑) | 0.261 | 0.279 | 0.274 | 0.215 | 0.219 | 0.222 | 0.254 | **0.358** | 28% | 0.340 | 0.313 | 0.337 | 0.336 | 0.395 |
| | | AV(↓) | 0.153 | 0.147 | 0.143 | 0.157 | 0.159 | 0.151 | 0.157 | **0.133** | 7% | 0.140 | 0.147 | 0.134 | 0.135 | 0.131 |
| | | SR(↑) | 1.710 | 1.902 | 1.921 | 1.377 | 1.380 | 1.477 | 1.623 | **2.680** | 39% | 2.435 | 2.126 | 2.523 | 2.495 | 3.011 |
| | | MDD(↓) | 0.098 | 0.117 | 0.106 | 0.193 | 0.223 | 0.185 | 0.166 | **0.072** | 26% | 0.074 | 0.083 | 0.094 | 0.078 | 0.072 |
| | | CR(↑) | 2.666 | 2.369 | 2.584 | 1.115 | 0.981 | 1.210 | 1.534 | **4.958** | 86% | 4.600 | 3.748 | 3.597 | 4.293 | 5.515 |

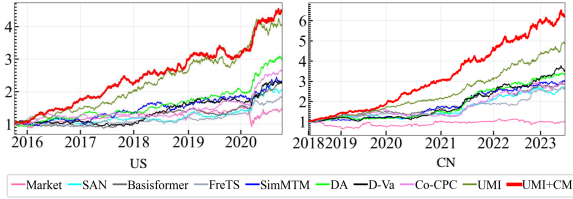

Figure 2: The Cumulative Wealth in US and CN markets.

Table 2: The universality of the irrationality factors.

| Data | Metrics | DA+F | Imp. | D-Va+F | Imp. | CPC+F | Imp. | *UMI* |
|---|---|---|---|---|---|---|---|---|
| US | AR(↑) | 0.241 | 10% | 0.233 | 32% | 0.249 | 17% | 0.289 |
| | AV(↓) | 0.153 | 12% | 0.149 | 5% | 0.153 | 2% | 0.144 |
| | SR(↑) | 1.580 | 25% | 1.570 | 38% | 1.628 | 19% | 2.007 |
| | MDD(↓) | 0.139 | 24% | 0.148 | 23% | 0.148 | 20% | 0.129 |
| | CR(↑) | 1.742 | 45% | 1.565 | 69% | 1.704 | 46% | 2.236 |
| Data | Metrics | DA+F | Imp. | D-Va+F | Imp. | CPC+F | Imp. | *UMI* |
| CN | AR(↑) | 0.342 | 31% | 0.348 | 25% | 0.334 | 22% | 0.358 |
| | AV(↓) | 0.142 | 7% | 0.139 | 5% | 0.137 | 4% | 0.133 |
| | SR(↑) | 2.411 | 41% | 2.502 | 32% | 2.443 | 27% | 2.680 |
| | MDD(↓) | 0.090 | 8% | 0.084 | 29% | 0.091 | 14% | 0.072 |
| | CR(↑) | 3.820 | 43% | 4.170 | 76% | 3.663 | 42% | 4.958 |

**Cross-Market Factor Learning Experiments.** The *UMI* model works as a factor-learning + forecasting-training framework. In the overall performance experiments, both stages occur within the same market. In fact, we can use one market data in the self-supervised learning process to enhance the performance of the other market. In the last column of Tab. 1 (*UMI* +CM), we conducted a cross-market factor learning experiment. For US market forecasting, we learned the market-level irrationality factor in the CN market (temporarily disabling the learnable ID representation $W_I$) and then further tuned the entire model in the US market. For CN market forecasting, we reversed the process. The cumulative wealth of *UMI* +CM is also shown in Fig. 2. The results show that *UMI* +CM has an improved performance, demonstrating our model's ability to leverage transferable knowledge across markets.

**Ablation Study.** Table 1 also shows the performance of four *UMI* model variants in the 2nd to 5th columns from the end: *i)* NS without the stock-level irrationality factors, *ii)* NM without the market-level irrationality factors, *iii)* NR without the RankIC regularization, and *iv)* ND without the Stock Relation Dependency module in Eq. (26) and Eq. (27), *i.e.,* only uses a Transformer as the forecasting function. The performance order is: NS < NM < ND < NR < UMI. These results highlight the importance of the modules to model performance. Additionally, stock-level factors seem more beneficial than the market-level, likely because they provide detailed, stock-specific information, while the market-level factors are uniform across all stocks.

**Universality of the Irrationality Factors.** The irrationality factors extracted by our model can also be used in various forecasting models. To evaluate this feature, we added these factors as inputs to three stock forecasting baselines (DoubleAdapt, D-Va, and Co-CPC). Table 2 shows the investment performance, with "+F" indicating models with additional factor inputs. The results demonstrate that

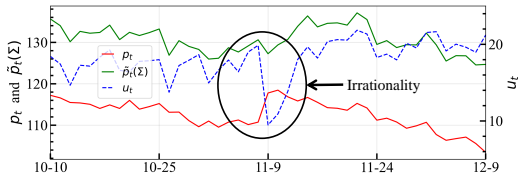

Figure 3: Visualization of $u_t$ value, cointegrated price $\tilde{p}_t$ ($\Sigma$) and price $p_t$ of stock SZ002253.

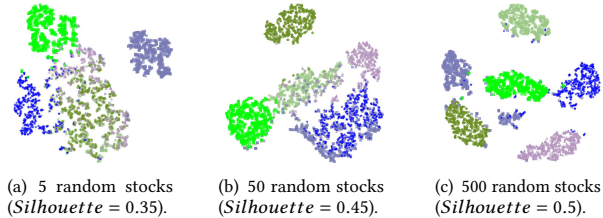

(a) 5 random stocks ($Silhouette = 0.35$).
(b) 50 random stocks ($Silhouette = 0.45$).
(c) 500 random stocks ($Silhouette = 0.5$).

Figure 4: Market representations from six days.

all models are significantly improved by our irrationality factors, confirming their universality. Despite these enhancements, our *UMI* model still outperforms the baselines, highlighting the effectiveness of our efficient forecasting model and RankIC regularization design.

### 6.3 Exploratory Analysis

Here, we illustrate how the irrationality factors work via visualized exploratory analysis.

**Stock-level Irrationality Factors.** Fig. 3 visualizes the series of the price $p_t$, estimated rational price $\tilde{p}_t$ ($\Sigma$) and the stock-level irrationality factors $u_t$ for the stock Wisesoft (SZ002253) in the CN market during 2016. Here, $p_t$ and $\tilde{p}_t$ ($\Sigma$) exhibit a cointegration relationship, with their difference, $u_t$, being stationary. There is an anomalous fluctuation of $u_t$ on Nov. 9, 2016, indicating an irrational event. This happened because the company's name is a homophone for "Trump's big win" and on that day, Donald Trump happened to be elected President of the United States, causing the stock price to hit its limit. This is a typical example of stock-level irrationality[6], demonstrating that our model can effectively capture irrational events in the market.

**Market-level Irrationality Factors.** Here, we examine how the individual stock dynamic representations construct the market-level irrationality factor. Specifically, we selected two trading days with the market-level irrational event of $\Delta_t = (1, 0, 0)$, two days with the event of $\Delta_t = (0, 1, 0)$, and two days without the irrational event. For each day, we generate market-level irrationality factors using a number of (5, 50, 500) randomly selected stocks. In Fig. 4(a), each point represents an instance of the market-level factor generated from 5 randomly selected stocks, with points of the same color corresponding to the same trading day. In Fig 4(b) and Fig 4(c), we increase the randomly selected stocks to 50 and 500. We used t-SNE [13] to convert the instances into 2D for visualization. We find that when the number of stocks reaches 500, days with different $\Delta_t$ can be clearly distinguished. This suggests our approach effectively identifies market-level irrational events from ordinary trading days. Additionally, days with the same $\Delta_t$ label are distinguishable by our factors, indicating that market-level factors provide sufficient information to capture overall market characteristics. The Silhouette score [39], which measures clustering cohesion, increases with the number of stocks, showing that a larger stock sample enhances the market representation capability of our factors.

## 7 Related Works

**Deep Learning for Stock Prediction.** In recent years, various types of deep neural network models have been introduced for stock data analysis [4, 11, 42, 45, 52, 54, 55, 59]. Some aim to forecast accurate stock prices/returns [53, 63], while others directly design trading strategies using reinforcement learning [8, 11, 31]. To better handle stock data, advanced deep learning methods proposed sophisticated network architectures [5, 15, 62]. Other methods incorporate supplementary information, such as fund manager details and stock comments, into the models [1, 27, 28]. Some self-supervised methods employ contrastive learning to exploit information in different types of positive/negative samples, such as daily-frequency and minutely-frequency [14], raw data and shuffled historical data [44], as well as correlated price movements [29].

**Stock Factor Learning.** For stock-level analysis, all deep learning-based methods can be considered as using representational learning to extract stock indicative factors [26, 30]. However, very few studies involve market irrationality. Moreover, in the stock-level factor analysis, we adopt a novel method to exploit cointegration relations, which is particularly innovative as this type of stock relationship is rarely considered. Most works focus on designing specialized neural networks [50, 51] to model stock relations, or incorporate additional information about stock relations, such as collective investments of funds [1, 28] and textual media about firm relevance [3]. For the market-level, traditional methods use stock market indexes to represent market overall state. A straightforward method is using the market indexes as a feature of stock forecasting models [58]. Some advanced methods use representations of some selected stocks, such as stocks with top-ranked returns, to indicate the overall state of market [51]. However, there are still few methods that focus on market irrationalities.

## 8 Conclusions

We introduce *UMI*, a method that exploit irrational events in stock markets from stock return forecasting. By pinpointing the gap between factual and estimated rational prices, *UMI* uncovers stock-level irrationalities. Moreover, it detects market-level irrationalities through representing anomalous synchronous of stocks. Extensive experiments in the U.S. and Chinese markets validated our model's exceptional performance and universality.

## Acknowledgments

This work is supported by the National Key R&D Program of China (2023YFC3304700). Prof. Jingyuan Wang's work was partially supported by the National Natural Science Foundation of China (No. 72222022, 72171013). Prof. Junjie Wu's work was partially supported by the National Natural Science Foundation of China (No. 72242101, 72031001), and the Outstanding Young Scientist Program of Beijing Universities (JWZQ20240201002).

---

[6]https://www.ft.com/content/22ffdb04-6d08-34e6-9543-73f744a500dc

# A Supplemental Materials

## A.1 Hedging Investment Strategy

We adopt a long-short hedging strategy as the backbone investment strategy for our forecasting model. We first give the concept of long/short position as follows.

DEFINITION 7 (LONG/SHORT POSITION). *The long position is the trading operation that buys a stock at time $t_1$ first and then sells it at $t_2$, while a short position is the trading operation that sells an stock at $t_1$ first and then buys it back at $t_2$.*

By taking a long position, traders expect a stock will rise in price, and the profit is $v^{(i)}(p_{t_2}^{(i)} - p_{t_1}^{(i)})$, where $v^{(i)}$ is the buying volume of the stock $s_i$. On the contrary, the trader's expectation in short position is that the price will drop, and the profit is $v^{(i)}(p_{t_1}^{(i)} - p_{t_2}^{(i)})$. In practice, a short position trader in the stock market borrows stocks from a broker and sells them at $t_1$, then at $t_2$, the trader buys the sold stocks back and returns them to the broker.

We then define the *long-short hedging* investment strategy adopted in our study, which uses long and short positions to hedge investment risks. Given a set of $M$ tradable stocks, the strategy first ranks the stocks in descending order by their returns in period $t$ as $s_t^{\text{rank}} = (s^{(i_1)}, \ldots, s^{(i_m)}, \ldots, s^{(i_M)})$, where $s^{(i_1)}$ is the largest return and $s^{(i_M)}$ is the smallest. The strategy takes long position on the stocks with top-$N$ returns, i.e., $s_t^{\text{long}} = (s^{(i_1)}, \ldots, s^{(i_N)})$ and short position on the stocks with bottom-$N$ returns, i.e., $s_t^{\text{short}} = (s^{(i_{M-N+1})}, \ldots, s^{(i_M)})$. Then the total return of the strategy in $t$ is simply the average of the returns of all traded stocks, which is given by

$$y_t^{\text{STR}} = \frac{1}{N}\left(\sum_{m=1}^{N} y_t^{(i_m)} - \sum_{m=M-N+1}^{M} y_t^{(i_m)}\right). \quad (33)$$

Note that without of generality we here assume all stocks are invested with the same amount.

## A.2 Forecasting Accuracy Metrics

• *Root Mean Square Error (RMSE)*: RMSE is defined as

$$\text{RMSE} = \sqrt{\frac{1}{T \times I} \sum_{t=1}^{T} \|\hat{\boldsymbol{y}}_t - \boldsymbol{y}_t\|_2^2}, \quad (34)$$

where $\|\cdot\|_2$ is the $l$-2 norm, $\boldsymbol{y}_t$ is the actual Rate of Return (RoR) vector at period $t$, $\hat{\boldsymbol{y}}_t$ is the predicted RoR and $I$ is the dimension of $\boldsymbol{y}_t$.

• *Mean Absolute Error (MAE)*: MAE is defined as

$$\text{MAE} = \frac{1}{T \times I} \sum_{t=1}^{T} |\hat{\boldsymbol{y}}_t - \boldsymbol{y}_t|, \quad (35)$$

where $|\cdot|$ is the $l$-1 norm.

• *Information Correlation (IC) and IC Information Ratio (ICIR)*: The Information Correlation between the actual RoR vector $\boldsymbol{y}_t$ and the predicted RoR vector $\hat{\boldsymbol{y}}_t$ is calculated by Eq. (30). We evaluate the prediction performance of our model using the average IC for all periods in the test set. Based on the IC, we define the IC Information Ratio (ICIR), which is obtained by dividing the average IC by its standard deviation.

• *Rank Information Correlation (RankIC) and RankIC Information Ratio (RankICIR)*: RankIC is computed by replacing $\hat{\boldsymbol{y}}$ and $\boldsymbol{y}$ with RoR rank vectors, i.e., $\hat{\boldsymbol{z}}$ and $\boldsymbol{z}$. Similarly, the RankICIR is obtained by dividing the average RankIC by its standard deviation. Compared to IC, RankIC ignores the numerical error of RoR, which only relates to the correctness of the ranking.

## A.3 Complete Training Procedure

Alg. 1 shows our complete training procedure. In *UMI*, we first define $\boldsymbol{p}_{t-1}^{(i)} = (p_1^{(i)}, \ldots, p_{t-1}^{(i)})$ and $\boldsymbol{P}_{t-1} = \{\boldsymbol{p}_{t-1}^{(i)}\}_{i=1}^{I}$. We then train the stock-level irrationality factor learning module using the cointegration objective functions $\mathcal{L}_S$. Subsequently, a sequence of stock-level irrationality factors is derived from a given sequence of stock prices, where

$$\boldsymbol{u}_{t-1}^{(i)}(\Sigma) = \left(u_1^{(i)}(\Sigma), \ldots, u_{t-1}^{(i)}(\Sigma)\right), \quad (36)$$

and

$$\boldsymbol{U}_{t-1}(\Sigma) = \left(\boldsymbol{u}_{t-1}^{(1)}(\Sigma), \ldots, \boldsymbol{u}_{t-1}^{(i)}(\Sigma), \ldots, \boldsymbol{u}_{t-1}^{(I)}(\Sigma)\right). \quad (37)$$

Next, the market-level irrational factor learning module is trained with the joint self-supervised market learning loss $\mathcal{L}_M = \mathcal{L}_C + \lambda_2 \mathcal{L}_P$. This allows us to derive a market-level irrational factor $\boldsymbol{m}_{t-1}$ from $\mathcal{E}_{t-1}$, which encompasses historical features of various stocks.

Lastly, the forecasting module is trained using the forecasting loss $\mathcal{L}$ as defined in Eq. (31). Subsequently, the forecasted stock return is then generated through a combination of a Transformer encoder, a stock relation dependency module and a MLP neural network.

**Algorithm 1** Factor-learning and Forecasting-training Procedure
---
Initializing parameters $B, W_C$ for stock-level irrationality factor learning.
**for** $episode = 1, 2, \ldots$ **do**
　　Optimizing loss $\mathcal{L}_S$ in training dataset.
**end for**
$U_{t-1} \leftarrow \text{Cointegration}(P_{t-1})$.
Initializing parameters $W_S, b, W_t, w_\eta, w_M, b_M$ for market-level irrational factor learning.
**for** $episode = 1, 2, \ldots$ **do**
　　Optimizing loss $\mathcal{L}_M = \mathcal{L}_C + \lambda_2 \mathcal{L}_P$ in training dataset.
**end for**
$m_{t-1} \leftarrow \text{Market}(\mathcal{E}_{t-1})$.
**for** $episode = 1, 2, \ldots$ **do**
　　Optimizing loss $\mathcal{L} = \mathcal{L}_{\text{MSE}} + \lambda_3 \mathcal{L}_{\text{RankIC}}$ in training dataset with stock-level irrationality factors $u$ and market-level irrationality factors $m$.
**end for**
$C_{t-1} \leftarrow \text{Transformer}(u_{t-1}, \mathcal{E}_{t-1})$.
$D_{t-1} \leftarrow \text{Dependency}(C_{t-1}, \mathcal{E}_{t-1})$.
$\hat{y}_t \leftarrow \text{MLP}(C_{t-1} \| D_{t-1} \| m_{t-1})$.
**return** $\hat{y}_t$

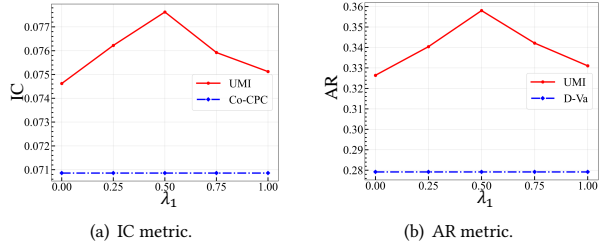

(a) IC metric.　　　(b) AR metric.

**Figure 5: Parameter sensitivity of $\lambda_1$ on CN markets for two metrics.**

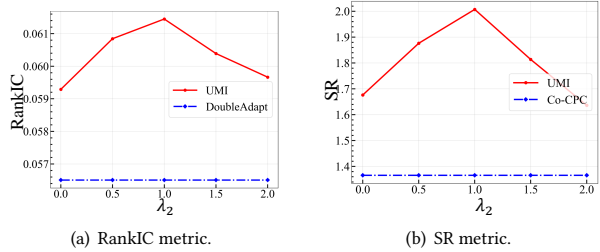

(a) RankIC metric.　　　(b) SR metric.

**Figure 6: Parameter sensitivity of $\lambda_2$ on US markets for two metrics.**

## A.4　Additional Universality of Irrationality Factors

Table 3 shows additional universality experiment results with forecasting performance. Analogous to Table 2, the forecasting performance of all models undergoes notable enhancement with the incorporation of the irrationality factors. This underscores the proficiency of our irrationality factors in harnessing valuable insights that may have been overlooked by the baselines.

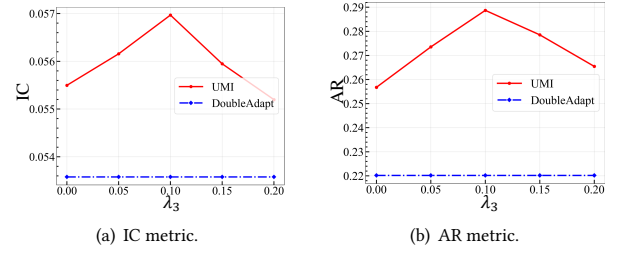

(a) IC metric.　　　(b) AR metric.

**Figure 7: Parameter sensitivity of $\lambda_3$ on US markets for two metrics.**

**Table 3: The universality of the irrationality factors with forecasting performance.**

| | | | US Market | | | |
|---|---|---|---|---|---|---|
| Metrics | DA+F | Imp. | D-Va+F | Imp. | CPC+F | Imp. | UMI |
| RMSE(↓) | .0395 | 1.0% | .0395 | 0.8% | .0394 | 1.6% | .0394 |
| MAE(↓) | .0231 | 1.2% | .0232 | 1.1% | .0232 | 0.8% | .0231 |
| IC(↑) | 0.056 | 4% | 0.054 | 11% | 0.054 | 4% | 0.057 |
| ICIR(↑) | 0.636 | 2% | 0.603 | 9% | 0.626 | 6% | 0.663 |
| RankIC(↑) | 0.057 | 2% | 0.058 | 4% | 0.062 | 14% | 0.061 |
| RankICIR(↑) | 0.642 | 1% | 0.637 | 5% | 0.697 | 16% | 0.666 |

| | | | CN Market | | | |
|---|---|---|---|---|---|---|
| Metrics | DA+F | Imp. | D-Va+F | Imp. | CPC+F | Imp. | UMI |
| RMSE(↓) | .0351 | 0.9% | .0352 | 1.2% | .0351 | 1.0% | .0350 |
| MAE(↓) | .0223 | 0.8% | .0223 | 1.0% | .0225 | 1.1% | .0222 |
| IC(↑) | 0.073 | 6% | 0.074 | 5% | 0.074 | 5% | 0.078 |
| ICIR(↑) | 0.776 | 6% | 0.798 | 7% | 0.803 | 5% | 0.831 |
| RankIC(↑) | 0.072 | 5% | 0.069 | 3% | 0.072 | 3% | 0.073 |
| RankICIR(↑) | 0.759 | 4% | 0.733 | 3% | 0.762 | 3% | 0.777 |

## A.5　Parameter Sensitivity

Fig. 5 reports the sensitivity of two evaluating metrics to the parameter $\lambda_1$ in Eq. (14) on the CN market. We also incorporate the best baseline methods on CN dataset for comparison. Here, we vary $\lambda_1$ in the set {0, 0.25, 0.5, 0.75, 1}. As shown in the figures, our model is relatively stable and consistently competitive with the best baseline methods.

Fig. 6 reports the sensitivity of two evaluating metrics to the parameter $\lambda_2$ in Eq. (24) on the US market. We also incorporate the best baseline methods on the US dataset for comparison. Here, we vary $\lambda_2$ in the set {0, 0.5, 1, 1.5, 2}.

Fig. 7 reports the sensitivity of two evaluating metrics to the parameter $\lambda_3$ in Eq. (31) on the US market. We also incorporate the best baseline methods on the US dataset for comparison. Here, we vary $\lambda_3$ in the set {0, 0.05, 0.1, 0.15, 0.2}. The sensitivities of other parameters are similar for the three parameters.